\documentclass[runningheads]{llncs}
\usepackage[T1]{fontenc}
\usepackage{graphicx}
\usepackage[table,xcdraw]{xcolor}
\usepackage{amsmath}
\usepackage{dirtree}
\usepackage{float} 
\usepackage{multirow}
\usepackage{subfig}
\usepackage{booktabs}

\begin{document}
%
\title{From Code to Field: Evaluating the Robustness of Convolutional Neural Networks for Disease Diagnosis in Mango Leaves}
\titlerunning{Evaluating CNN Robustness in Mango Leaf Disease Detection}
%
%

\author{Gabriel Vitorino de Andrade\inst{1}\orcidID{0009-0005-6130-8193} \and Saulo Roberto dos Santos\inst{1}\orcidID{0009-0007-8520-8124} \and Itallo Patrick Castro Alves da Silva\inst{1}\orcidID{0009-0008-8543-7776} \and Emanuel Adler Medeiros Pereira\inst{2}\orcidID{0000-0002-6694-5336}\and Erick de Andrade Barboza\inst{1}\orcidID{0000-0002-0558-9120}}

%

\authorrunning{G. V. de Andrade et al.}
%
\institute{Instituto de Computação, Universidade Federal de Alagoas, Maceió, AL, 57072-970, Brazil\\
\email{\{gva,srs,ipcas,erick\}@ic.ufal.br}\\
\and
Centro de Tecnologia, Universidade Federal do Rio Grande do Norte, Natal, RN, 59078-900, Brazil\\
\email{emanuel.pereira.111@ufrn.edu.br}}
\maketitle              
\begin{abstract}
The validation and verification of artificial intelligence (AI) models through robustness assessment are essential to guarantee the reliable performance of intelligent systems facing real-world challenges, such as image corruptions including noise, blurring, and weather variations. Despite the global importance of mango (Mangifera indica L.), there is a lack of studies on the robustness of models for the diagnosis of disease in its leaves. This paper proposes a methodology to evaluate convolutional neural networks (CNNs) under adverse conditions. We adapted the MangoLeafDB dataset, generating MangoLeafDB-C with 19 types of artificial corruptions at five severity levels. We conducted a benchmark comparing five architectures: ResNet-50, ResNet-101, VGG-16, Xception, and LCNN (the latter being a lightweight architecture designed specifically for mango leaf diagnosis). The metrics include the F1 score, the corruption error (CE) and the relative mean corruption error (relative mCE). 
The results show that LCNN outperformed complex models in corruptions that can be present in real-world scenarios such as Defocus Blur, Motion Blur, while also achieving the lowest mCE. Modern architectures (e.g., ResNet-101) exhibited significant performance degradation in corrupted scenarios, despite their high accuracy under ideal conditions. These findings suggest that lightweight and specialized models may be more suitable for real-world applications in edge devices, where robustness and efficiency are critical. The study highlights the need to incorporate robustness assessments in the development of intelligent systems for agriculture, particularly in regions with technological limitations.


\end{abstract}

\section{Introduction}

Deep neural networks and machine learning techniques have been widely used in various computer vision tasks, such as object classification. 
However, unlike humans who can deal with different changes in image structures and styles such as snow, \emph{blur} and pixelation, computer vision models cannot differentiate in the same way \cite{hendrycks2018benchmarking}. 
As a result, the performance of neural networks declines when the images used as input for the model are affected by natural distortions. This highlights the need for system validation and verification to ensure that models perform as expected under different conditions. 
In production settings, where models will inevitably encounter distorted inputs \cite{trinh2024improvingrobustnesscorruptionsmultiplicative}, ensuring thorough system validation and verification processes is crucial. 
For example, autonomous vehicles must be able to cope with extremely variable external conditions, such as fog, frost, snow, sandstorms, or falling leaves. It is impossible to predict all potential conditions that can occur in nature \cite{michaelis2019whenwinteriscoming}.

Because of this, achieving the kind of robustness that humans possess is an important goal for computer vision and machine learning, as well as creating models that can be deployed in safety-critical applications \cite{hendrycks2018benchmarking}. 
Therefore, robust system verification and validation become essential in ensuring that these systems perform reliably. 
The robustness of models against different types of perturbation has been a much-studied topic in the machine learning community \cite{croce2021robustbench}. 
Natural corruptions, which are an important type of disturbance \cite{croce2021robustbench}, are common in real scenarios and can reduce the accuracy of models \cite{hendrycks2018benchmarking}, so their study, in conjunction with the validation and verification processes of the system, has been widely carried out \cite{hendrycks2018benchmarking,croce2021robustbench,Hendrycks_2021_ICCV}.

In parallel, modern technologies, including machine learning and computer vision, have been increasingly applied to agriculture to enhance productivity and sustainability~\cite{modern2016agriculture}. These techniques have introduced innovative trends in monitoring and forecasting~\cite{ml2021agriculturereview}, which contribute directly to agricultural improvements~\cite{modern2016agriculture}. Machine learning models have shown great potential to detect diseases in crop leaves~\cite{detect2023mangodisease}, a critical task given that pests and diseases affect an estimated 40\% of food crops globally~\cite{Citaristi2022fao}. Among economically important crops, mango (\textit{Mangifera indica L.}) ranks as the fifth most cultivated fruit worldwide~\cite{fao2020}, which thrives particularly in tropical and subtropical regions~\cite{mangotree2021}.

Given the importance of reliable detection systems in agriculture, robustness is particularly important in this context because diagnostic systems are expected to operate under real-world conditions - including mobile or edge devices - where image capture is subject to noise, blur, and lighting variability.

The research by \cite{detect2023mangodisease} introduces neural network models to classify leaf diseases of plants, describing their classification performance. 
Furthermore, \cite{mango-2024-tflite} provides a model for the classification of mango leaf disease on mobile devices. 
However, these studies do not assess the robustness of the model to corruption. 
In contrast, works such as \cite{hendrycks2018benchmarking} and \cite{croce2021robustbench} extensively analyze the robustness of cutting-edge computer vision models, but do not address the detection of mango leaf disease.

In this work, our aim is to bridge this gap by proposing a methodology to evaluate the robustness of CNN models in the task of determining mango leaf disease. 
To this end, we introduce MangoLeafDB-C, a corrupted version of the MangoLeafDB dataset \cite{detect2023mangodisease} that incorporates 19 types of synthetic distortions at five severity levels \cite{hendrycks2018benchmarking}. 
We conducted a benchmark study across five CNN architectures: ResNet-50, ResNet-101, VGG-16, Xception, and LCNN - a lightweight network tailored for the detection of mango leaf disease.

The work is organized as follows.
In Section \ref{sec:related-work}, we present the related works in relation to our study.
In Section \ref{sec:methodology}, we present in detail the proposed methodology, including the construction of MangoLeafDB-C and the evaluation protocols.
In Section \ref{sec:results}, we present and analyze the experimental results.
In Section \ref{sec:conclusion}, we give some conclusions, comments, and ideas for future work.


\section{Related Work}
\label{sec:related-work}

The study presented in \cite{detect2023mangodisease} proposed a lightweight convolutional neural network (LCNN) to diagnose seven distinct mango leaf diseases in Bangladesh. 
The study used the MangoLeafDB dataset \cite{ali2022mangoleafbd}, which contains 4,000 images classified into eight categories, including diseased and healthy leaves. The LCNN model was compared to pre-trained architectures such as VGG16, ResNet50, ResNet101, and Xception, achieving the highest test accuracy of 98\%. 

The study presented in \cite{hendrycks2018benchmarking} established a rigorous \emph{benchmark} for robustness in image classifiers. 
To this end, datasets such as IMAGENET-C and IMAGENET-P were created. While IMAGENET-C standardized and expanded the topic of robustness against corruption, IMAGENET-P allows researchers to evaluate the robustness of a classifier against common perturbations. 
The idea of this \emph{benchmark} is to evaluate the performance of models against common corruptions (IMAGENET-C) and perturbations (IMAGENET-P). 
The paper also defines robustness to corruption and disturbance and differentiates them from robustness to adversarial disturbances. Finally, smaller datasets were created with the same purpose as IMAGENET-C, such as: CIFAR-10-C, CIFAR-100-C, TINY IMAGENET-C, and IMAGENET 64 X 64-C. 
The metrics \emph{Mean Corruption Error} (mCE) and \emph{Relative Mean Corruption Error} (\emph{Relative} mCE) were proposed and used to evaluate the robustness against corruption.

The study presented in \cite{croce2021robustbench} established a standardized reference on adversarial robustness in neural network models. 
To do this, it used the task of image classification. 
With this in mind, the idea of the work was to establish a real tracking of the progress of studies on adversarial robustness in the literature. The \emph{benchmark} evaluated common corruptions\cite{hendrycks2018benchmarking}, \(\ell_\infty\)- and \(\ell_2\)-robustness. 
In addition, \emph{AutoAtack}\cite{pmlr-v119-croce20b} was used to standardize the robustness assessment of \(\ell_p\) and CIFAR-10-C\cite{hendrycks2018benchmarking} for the robustness assessment against common corruptions. 
Finally, a platform has been made available with more than 120 evaluated models and aims to reflect the state-of-the-art in evaluating the robustness of models in image classification tasks.


\section{Methodology}
\label{sec:methodology}

We follow a methodology that can be divided into three main steps. 
Initially, we created a corrupted version of the original MangoLeafDB dataset. 
Then, we implemented and validated the five CNN models used in \cite{detect2023mangodisease}.
Finally, we calculate robustness metrics considering the corrupted dataset and the CNN models following the methodology and metrics proposed in \cite{hendrycks2018benchmarking}. 
The purpose of this methodological approach is to allow evaluation of the model performance under different image degradation conditions, simulating real application scenarios, and offering valuable insights into their robustness.

\subsection{Tools and Dependencies Used}
The pipeline was constructed in Python 3.9.13 on Windows, using pip for package management. 
We created the synthetic MangoLeafDB-C database using scipy (1.13.1), wand (0.6.13) and ImageMagick (7.1.1-47). 
The operations on the images and tensors involved libraries such as Pillow (11.1.0), torch (2.6.0), torchvision (0.21.0), numpy (1.26.4) and scikit-image (0.24.0). 
Our CNN models (ResNet50, ResNet101, VGG-16, Xception, LCNN) were trained and evaluated using TensorFlow (2.10.0) and Keras (2.10.0), with h5py (3.13.0) for model persistence. 
OpenCV (4.11.0) served for image preprocessing. For data splitting and metric calculations such as accuracy\_score, scikit-learn (1.6.1) was used, while pandas (2.2.3), matplotlib (3.9.4), and Plotly (6.0.0) were used for data analysis and visualization. 
All dependencies are specified by version, and our code is publicly available for reproducibility
\footnote{\url{https://github.com/GabrielKcin900/research-repository.git}}


\subsection{MangoLeafDB-C Creation}
\label{subsec:mangoleafDB-C subsection}

To assess the robustness of the models for classifying mango leaf diseases in the presence of common digital corruptions, it was necessary to create a corrupted version of the original dataset. 
We call this new version MangoLeafDB-C.

The methodology for creating MangoLeafDB-C was directly inspired by the procedure used to build the ImageNet-C dataset, proposed in \cite{hendrycks2018benchmarking}.
This methodology considers 19 different types of digital corruption: Brightness, Contrast, Defocus Blur, Elastic, Fog, Frost, Gaussian Blur, Glass Blur, Impulse Noise, JPEG, Motion Blur, Pixelate, Saturate, Shot Noise, Snow, Spatter, Speckle Noise, and Zoom Blur.

\begin{figure}[tbp]
  \centering
  \includegraphics[width=0.8\textwidth]{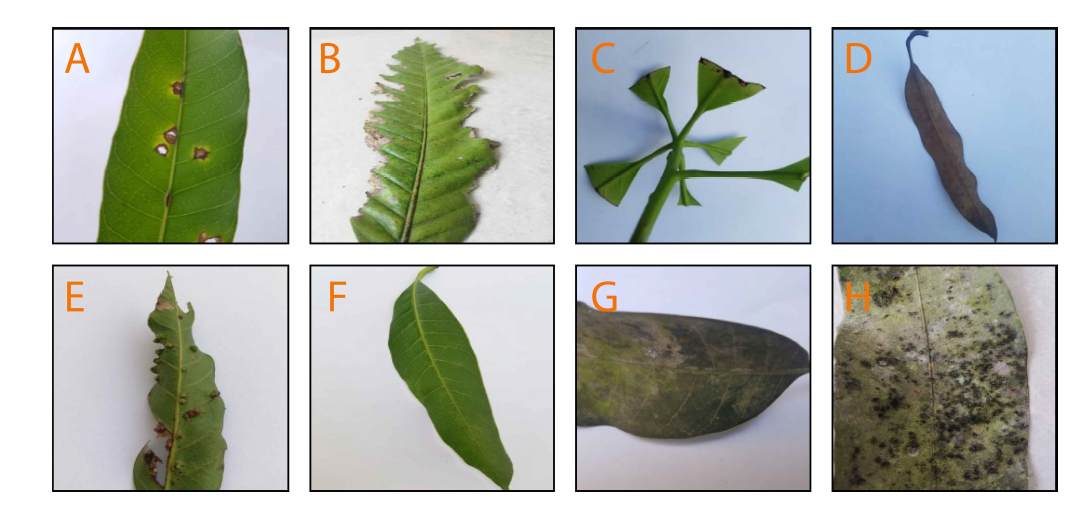}
  \caption{Sample of Mango leaf diseases : A) Anthracnose B) Bacterial Canker C) Cutting Weevil D) Die Back E) Gall Midge F) Healthy G) Powdery Mildew H) Sooty Mould. Source: \cite{detect2023mangodisease}}
  \label{fig:mangoleafdb sample}
\end{figure} 

The starting point for the creation of MangoLeafDB-C was the MangoLeafDB dataset, publicly available on Kaggle\footnote{\url{https://www.kaggle.com/datasets/aryashah2k/mango-leaf-disease-dataset/data}}. 
This data set consists of 4,000 images of mango leaves, classified into eight distinct classes of disease (Anthracnose, Bacterial Canker, Cutting Weevil, Die Back, Gall Midge, Healthy, Powdery Mildew and Sooty Mold) or healthy condition. 
Each class contains 500 images. 
Figure \ref{fig:mangoleafdb sample} shows one sample of images for each disease.
We use this specific dataset to ensure a direct comparison with the work of \cite{detect2023mangodisease}. 



\begin{figure}[tbp]
  \centering
  \includegraphics[width=0.9\textwidth]{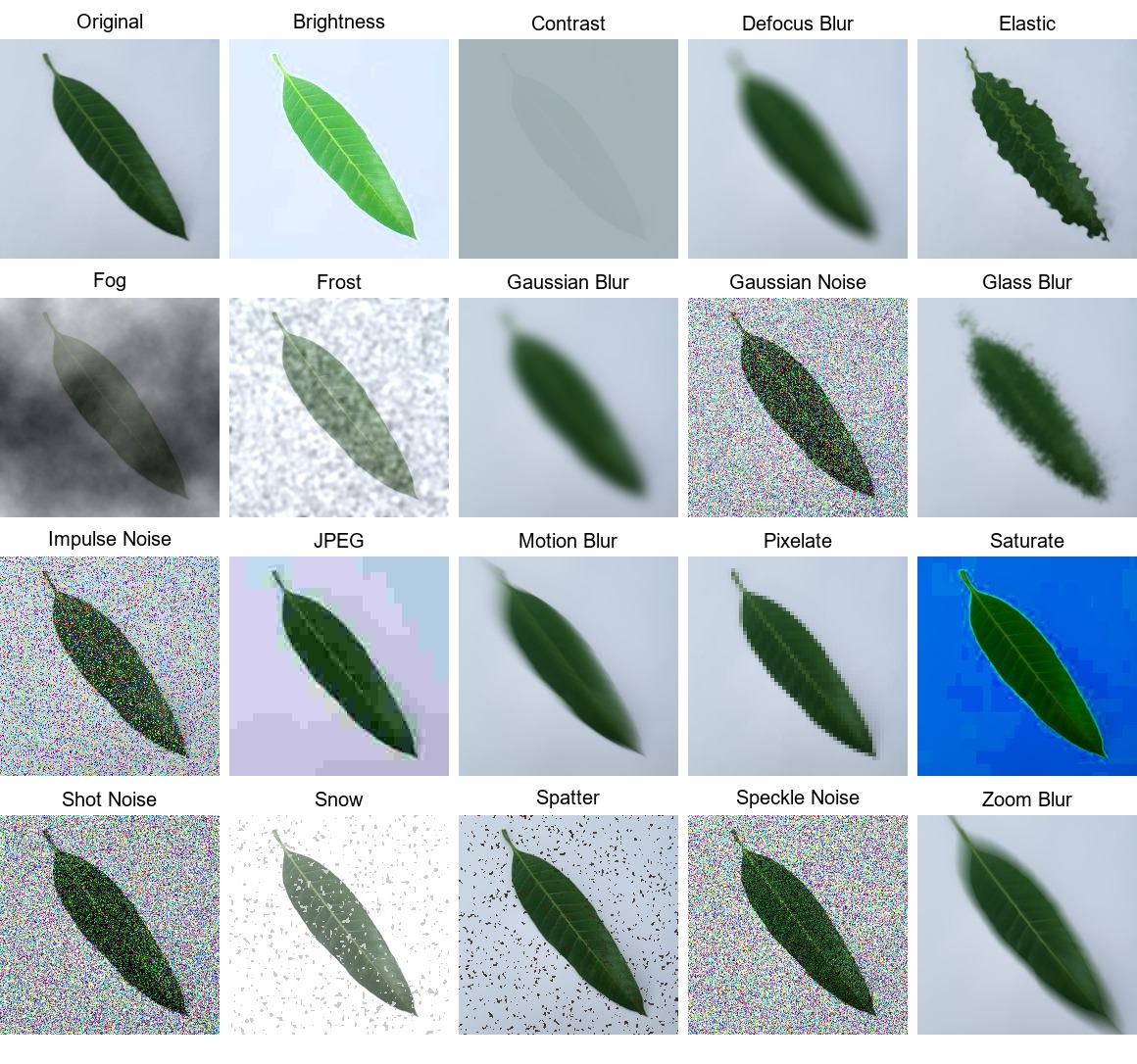}
  \caption{Image of a healthy mango leaf in its original format and after the application of the 19 corruptions considered in this work with the highest severity level.}
  \label{fig:demonstration_corrupted-bank}
\end{figure}

To apply corruptions to MangoLeafDB, we adapt the script\footnote{\url{https://github.com/hendrycks/robustness/tree/master/ImageNet-C}} shared by the authors of \cite{hendrycks2018benchmarking} to process the specific images and directory structure of MangoLeafDB.
For each type of corruption, five different levels of severity were defined, ranging from 1 (lowest intensity of corruption) to 5 (highest intensity). Applying each corruption at each severity level to all 4,000 images in the original MangoLeafDB generated 95 subsets of corrupted data. 
Each subset preserves the original structure of the MangoLeafDB classes, composed of 8 classes with 500 images each. 
Figure \ref{fig:demonstration_corrupted-bank} shows a sample of an image of a healthy mango leaf with all corruptions applied with severity level 5.

\subsection{CNNs Implementation and Validation}
\label{subsec:cnn-implementation-validation subsection}

This part of the methodology is a replication of the methodology proposed in \cite{detect2023mangodisease}, which evaluates five CNN models: ResNet50, ResNet101, VGG16, Xception, and a LCNN (Lightweight Convolutional Neural Network) proposed by the authors for the classification of diseases in mango leaves. 
In \cite{detect2023mangodisease}, all models (except LCNN) involve transfer learning, but the details and codes of the architecture are not disclosed. 
However, the authors provide the classification metric (precision, recall, F1-Score) for each class, allowing for comparative validation. 
Next, we describe our implementation strategy to ensure equivalence with the original study.

To replicate the experiments, we strictly followed the hyperparameters described in \cite{detect2023mangodisease}.
We set the epoch count to 50, batch size to 32, and learning rate at 0.001, utilizing categorical cross entropy as the loss function. The data was split with an 80:10:10 ratio for training, validation, and testing.
All models (except LCNN) were implemented using TensorFlow/Keras. 
We adapted the default Keras implementations by replacing the final classification layer (originally fine-tuned for 1,000 classes of ImageNet) with a dense layer with 8 units (softmax) for the 8 classes of the problem. 
For LCNN, we reconstructed the architecture as described in \cite{detect2023mangodisease}.


We standardized images to 224×224 pixels across RGB channels for all models, used pre-trained ImageNet weights for transfer learning models, and utilized Adam optimizer with a constant learning rate of 0.001.

The validation involved matching our implementation's classification metrics with those of \cite{detect2023mangodisease}. 
For transfer learning models, we iteratively refined final layers (e.g., pooling strategies, adding dense layers, Flatten) to minimize differences. 
The structure of the LCNN architecture was validated by comparing convolutional blocks and activation functions with the original blueprint.
Table \ref{tab:model_validation} presents the F1 score of our implementation alongside that of \cite{detect2023mangodisease}.
The F1 scores obtained in this work show a similarity of 97.12\% with those reported in \cite{detect2023mangodisease}, with an average difference of only 2. 88\%, which confirms the fidelity of our implementation.




\begin{table}[htb]
\centering
\begin{tabular}{cccc}
\hline
Model     & F1-score & F1-score & Accuracy \\ 
     & \cite{detect2023mangodisease} & [this Work] & [this work] \\ 
\hline
ResNet50  & 0.61                                                   & 0.66   & 0.67 \\
ResNet101 & 0.68                                                   & 0.68   & 0.68 \\
VGG16     & 0.97                                                   & 0.95   & 0.95 \\
Xception  & 0.96                                                   & 0.93   & 0.94 \\
LCNN      & 0.98                                                   & 0.97   & 0.97 \\ \hline
\end{tabular}
\caption{The F1 score reported in \cite{detect2023mangodisease}, the F1 score, and the accuracy on the test dataset obtained in this work.}
  \label{tab:model_validation}
\end{table}


\subsection{Robustness Evaluation}
\label{subsec:robustness-evaluation subsection}

Following the benchmark protocol established by \cite{hendrycks2018benchmarking}, the CNN models were trained exclusively on the clean dataset (MangoLeafDB).
The robustness evaluation was performed considering the corrupt dataset (MangoLeafDB-C) as a test data.

We first produced classification reports for each CNN model across all corruption types (c) and severity levels (s, 1 to 5), which offered class-specific performance metrics, including the F1-score. 
From these reports, we derived the average F1 score for each pair (c, s).

Then, we conducted a comprehensive analysis of the impacts of corruption, identifying those with minimal and significant effects on network performance, and ranking the models by F1 score for each corruption. 
Additionally, we examined model performance variability based on corruption ranks, offering insight into each model's robustness distribution against various degradations.

To evaluate robustness, we used the metrics introduced in \cite{hendrycks2018benchmarking}, specifically the Corruption Error (CE) and the relative Corruption Error (Relative CE). 
CE assesses classifier performance under a certain corruption, normalized against a reference model's performance in the same scenario. 
The formula for computing CE for a corruption \(c\) is presented in Equation \ref{eq:CE}.

\begin{equation}
    CE^f_c = (\sum_{s=1}^{5} E^f_{s,c}) / (\sum_{s=1}^{5} E^{ResNet101}_{s,c})
    \label{eq:CE}
\end{equation}
where $ E^f_{s,c} $ is the top-1 error rate of the classifier $f$ for corruption $c$ at severity level $s$, and $ E^{ResNet101}_{s,c} $ is the top-1 error rate of ResNet101 for the same corruption and severity. 
Adopting the methodology from \cite{detect2023mangodisease}, we selected ResNet101 as the normalization reference, given its lowest accuracy (highest error) among the models evaluated considering the clean dataset, as shown in Table \ref{tab:table_mce}. (Note: To calculate mCE and relative mCE, we considered a test set different from that used to create Table \ref{tab:model_validation}.)

Relative CE evaluates the performance degradation of a classifier $f$ under corruption $c$ relative to its performance on the clean dataset, compared to the degradation of the reference model (ResNet101) as shown in Equation \ref{eq:relative_CE}.

\begin{equation}
    Relative~{CE}^f_c = \sum_{s=1}^{5}(E^f_{s,c} - E^f_{clean}) / \sum_{s=1}^{5}(E^{ResNet101}_{s,c} - E^{ResNet101}_{clean})
    \label{eq:relative_CE}
\end{equation}
where $E^f_{clean}$ and $E^{ResNet101}_{clean}$ are the top-1 error rates of classifier $f$ and ResNet101, respectively, on the clean dataset. 
This metric captures the gap between performance on clean and corrupted data, relativized by the degradation of the reference model.

To thoroughly evaluate the model's robustness, we utilize mean CE (mCE) and Relative mean CE (Relative mCE). The mCE is the average CE for all 19 corruptions in MangoLeafDB-C, while the relative mCE represents the average relative CE for these corruptions. 
Lower mCE values indicate higher robustness. 
Relative mCE assesses the overall relative robustness, reflecting performance degradation due to corruption. 


\section{Results and Discussion}
\label{sec:results}

In this section, we first detail the F1 score trends, then quantify model-specific corruption sensitivities through rankings, and finally examine the trade-offs between clean accuracy and robustness using corruption error metrics.

\subsection{F1–Score Degradation Patterns}
\begin{figure}[tbp]
  \centering
  \includegraphics[width=\textwidth]{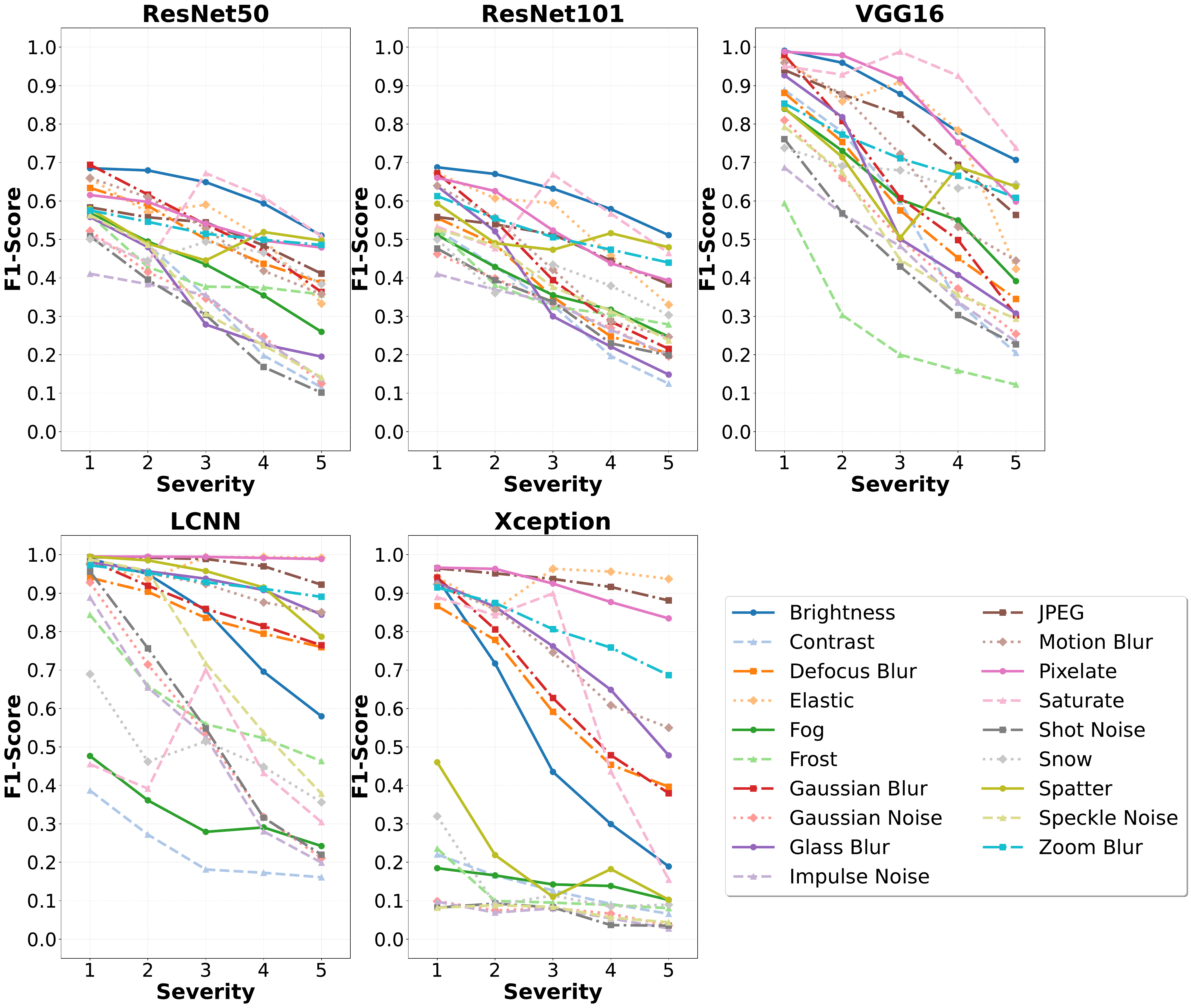}
  \caption{Macro F1–Score vs.\ corruption severity considering the corruptions defined in \cite{hendrycks2018benchmarking} and the CNN models defined in \cite{detect2023mangodisease}.}
  \label{fig:f1_all}
\end{figure}

Figure~\ref{fig:f1_all} illustrates the macro-averaged F1 score as a function of the severity of corruption. 
Several important observations emerge from the results. 
A consistent trend across all models is the overall decrease in the F1 score as the severity of corruption increases; however, the rate and magnitude of this decline vary considerably depending on the specific architecture and the nature of the corruption.

Examining the performance of individual architectures reveals distinct robustness profiles. 
For example, ResNet-50 and ResNet-101 demonstrate relatively stable performance at lower severity levels for various types of corruption, showing a more gradual decline compared to VGG-16 and Xception, which often exhibit a steeper drop in the F1 score even at moderate levels of corruption. 
In contrast, for most corruptions, ResNet-50 and ResNet-101 returned the worst F1 score with a lower severity ($s=1$)


In particular, Xception and LCNN exhibit good robustness, with much flatter performance curves under geometric and compression distortions. LCNN, in particular, maintains an F1 score above 0.9 on Pixelate and Elastic at all severity levels, indicating its strong ability to capture shape-based features even under distortion.

Finally, all models exhibit significant vulnerability to random noise corruptions such as Impulse, Speckle, and Shot. At the highest severity level ($s=5$), F1 scores on noises such as Contrast, Fog, Frost, Gaussian Noise, Impulse Noise, Shot Noise, and Speckle Noise fall below 0.4631 for all models, emphasizing the need for robustness strategies specific to these noises in future research.


Table~\ref{tab:corruption_ranking} presents a ranked list of all 19 types of corruption for each model, ordered by the averaged F1 score, from highest (rank 1, least damaging) to lowest (rank 19, most damaging). 
Some important observations can be drawn from these data.
First, the least damaging distortions such as Elastic, Pixelate, and Zoom Blur tend to occupy the top positions across all evaluated networks. 
This suggests that transformations that preserve the global structure of leaves tend to have a minimal impact on classification performance.

Second, there is a notable divergence in sensitivity to specific corruptions across architectures. 
While ResNet models identify Impulse Noise and Shot Noise as the most damaging, Xception and LCNN models are more severely degraded by Impulse Noise and Contrast, respectively. 
These differences point to architecture-specific biases in feature extraction and robustness.

The disparity in F1 performance between minimal and maximal corruptions is significant. 
For deeper models like ResNet-50 and ResNet-101, their F1 score decreases approximately by 0.3, dropping from 0.6160 to 0.3161 and 0.6235 to 0.2953, respectively. 
In contrast, LCNN exhibits an even greater change, from 0.9930 to 0.2347. 
Despite the varied proportions of F1 score decline, it is important to acknowledge the substantial difference in the maximum F1 scores.

\begin{table}[tbp]
  \centering
    \renewcommand{\arraystretch}{1.2} 
  \resizebox{\textwidth}{!}{%
    \begin{tabular}{c|l|l|l|l|l|}
      \cline{2-6}
      & \textbf{ResNet-50} & \textbf{ResNet-101} & \textbf{VGG-16} & \textbf{Xception} & \textbf{LCNN} \\
      \hline
      \multicolumn{1}{|c|}{\textbf{Rank 1}}  & \cellcolor[HTML]{9AFF99}Brightness (0.6160) & \cellcolor[HTML]{9AFF99}Brightness (0.6235) & \cellcolor[HTML]{9AFF99}Saturate (0.9064) & \cellcolor[HTML]{9AFF99}Elastic (0.9308) & \cellcolor[HTML]{9AFF99}Pixelate (0.9930) \\
      \hline
      \multicolumn{1}{|c|}{\textbf{Rank 2}}  & \cellcolor[HTML]{9AFF99}Saturate (0.5423)   & \cellcolor[HTML]{9AFF99}Saturate (0.5480)   & \cellcolor[HTML]{9AFF99}Brightness (0.8630) & \cellcolor[HTML]{9AFF99}JPEG (0.9299)   & \cellcolor[HTML]{9AFF99}Elastic (0.9789)  \\
      \hline
      \multicolumn{1}{|c|}{\textbf{Rank 3}}  & \cellcolor[HTML]{9AFF99}Elastic (0.5323)    & \cellcolor[HTML]{9AFF99}Pixelate (0.5461)   & \cellcolor[HTML]{9AFF99}Pixelate (0.8468)   & \cellcolor[HTML]{9AFF99}Pixelate (0.9131)  & \cellcolor[HTML]{9AFF99}JPEG (0.9732)     \\
      \hline
      \multicolumn{1}{|c|}{\textbf{Rank 4}}  & Pixelate (0.5280)    & Gaussian Blur (0.5364)  & Elastic (0.7893)       & Zoom Blur (0.8080)     & Zoom Blur (0.9312)   \\
      \hline
      \multicolumn{1}{|c|}{\textbf{Rank 5}}  & Zoom Blur (0.5173)   & Elastic (0.5307)        & JPEG (0.7801)          & Motion Blur (0.7369)   & Spatter (0.9279)     \\
      \hline
      \multicolumn{1}{|c|}{\textbf{Rank 6}}  & Spatter (0.5104)     & Zoom Blur (0.5241)      & Zoom Blur (0.7221)     & Glass Blur (0.7359)    & Glass Blur (0.9256)  \\
      \hline
      \multicolumn{1}{|c|}{\textbf{Rank 7}}  & JPEG (0.4878)        & JPEG (0.5162)           & Motion Blur (0.7074)   & Gaussian Blur (0.6464) & Motion Blur (0.9153) \\
      \hline
      \multicolumn{1}{|c|}{\textbf{Rank 8}}  & Motion Blur (0.4301) & Motion Blur (0.5148)    & Spatter (0.6772)       & Saturate (0.6445)      & Gaussian Blur (0.8684) \\
      \hline
      \multicolumn{1}{|c|}{\textbf{Rank 9}}  & Gaussian Blur (0.4233)& Defocus Blur (0.5093)   & Snow (0.6769)          & Defocus Blur (0.6171)  & Defocus Blur (0.8466) \\
      \hline
      \multicolumn{1}{|c|}{\textbf{Rank 10}} & Snow (0.3956)        & Spatter (0.5061)        & Gaussian Blur (0.6393) & Brightness (0.5157)    & Brightness (0.8139)  \\
      \hline
      \multicolumn{1}{|c|}{\textbf{Rank 11}} & Speckle Noise (0.3877)& Snow (0.4579)          & Fog (0.6229)           & Spatter (0.2147)       & Speckle Noise (0.7157)
      \\
      \hline
      \multicolumn{1}{|c|}{\textbf{Rank 12}} & Fog (0.3718)         & Fog (0.4224)            & Defocus Blur (0.6013)  & Fog (0.1467)           & Frost (0.6097)       \\
      \hline
      \multicolumn{1}{|c|}{\textbf{Rank 13}} & Defocus Blur (0.3687)& Frost (0.4201)          & Glass Blur (0.5922)    & Snow (0.1396)          & Shot Noise (0.5593)  \\
      \hline
      \multicolumn{1}{|c|}{\textbf{Rank 14}} & Glass Blur (0.3657)  & Glass Blur (0.3479)     & Contrast (0.5613)      & Contrast (0.1341)      & Gaussian Noise (0.5408)
      \\
      \hline
      \multicolumn{1}{|c|}{\textbf{Rank 15}} & Frost (0.3634)       & Speckle Noise (0.3443)  & Gaussian Noise (0.5195)& Frost (0.1200)         & Impulse Noise (0.5098)
      \\
      \hline
      \multicolumn{1}{|c|}{\textbf{Rank 16}} & Gaussian Noise (0.3321)& Contrast (0.3435)     & Speckle Noise (0.5129)& Gaussian Noise (0.0712) & Snow (0.4940)       \\
      \hline
      \multicolumn{1}{|c|}{\textbf{Rank 17}} & \cellcolor[HTML]{FFCCC9}Shot Noise (0.3272) & \cellcolor[HTML]{FFCCC9}Gaussian Noise (0.3311) & \cellcolor[HTML]{FFCCC9}Impulse Noise (0.4616) & \cellcolor[HTML]{FFCCC9}Speckle Noise (0.0709) & \cellcolor[HTML]{FFCCC9}Saturate (0.4567) \\
      \hline
      \multicolumn{1}{|c|}{\textbf{Rank 18}} & \cellcolor[HTML]{FFCCC9}Contrast (0.3209) & \cellcolor[HTML]{FFCCC9}Impulse Noise (0.3053) & \cellcolor[HTML]{FFCCC9}Shot Noise (0.4577) & \cellcolor[HTML]{FFCCC9}Shot Noise (0.0658) & \cellcolor[HTML]{FFCCC9}Fog (0.3299) \\
      \hline
      \multicolumn{1}{|c|}{\textbf{Rank 19}} & \cellcolor[HTML]{FFCCC9}Impulse Noise (0.3161) & \cellcolor[HTML]{FFCCC9}Shot Noise (0.2953) & \cellcolor[HTML]{FFCCC9}Frost (0.2757) & \cellcolor[HTML]{FFCCC9}Impulse Noise (0.0657) & \cellcolor[HTML]{FFCCC9}Contrast (0.2347)
      \\
      \hline
    \end{tabular}%
  }
  \caption{Ranking of all 19 corruption types per model (Rank 1 = least damaging, Rank 19 = most damaging), based on average macro F1 across severity levels.}
  \label{tab:corruption_ranking}
\end{table}

\subsection{Corruption Error Metrics}

Tables~\ref{tab:table_mce} and~\ref{tab:relative_mce} report the Clean Error and Mean Corruption Error (mCE), along with the relative error metrics per corruption, for all models normalized to ResNet-101. Table~\ref{tab:table_mce} shows the Clean Error, the absolute mCE (set to 100 for ResNet-101) and the CE per corruption.
Table~\ref{tab:relative_mce} presents the mean relative corruption error (rel. mCE) and the relative CE per corruption (with rel. mCE = 100 for ResNet-101). 
The results reveal that while ResNet-101 and ResNet-50 achieve similar overall robustness, the shallower ResNet-50 is notably more sensitive to blur and certain digital corruptions. 
Xception, despite a lower absolute mCE than ResNets, is particularly vulnerable/brittle to noise-based corruptions, with relative CEs above 130 for several types of noise. 
LCNN achieves the lowest overall absolute mCE of the general models, performing particularly well on blur and digital distortions, but is less robust to weather-related corruptions such as frost and fog.

LCNN demonstrates superior robustness, as evidenced by the lowest absolute mCE and consistently minimal error rates across most types of corruption. 
Specifically, LCNN achieved the highest F1 score for 14 of 19 corruptions, including Defocus Blur, Elastic, Frost, Gaussian Blur, Gaussian Noise, Glass Blur, Impulse Noise, JPEG, Motion Blur, Pixelate, Shot Noise, Spatter, Speckle Noise, and Zoom Blur. 
It secured the second-highest F1 score for Brightness and Snow and ranked fourth for Contrast and Fog, with the lowest F1 score observed for Saturate corruption.


\begin{table}[tbp]
\centering
\begin{minipage}{0.65\textwidth}
  \centering
  \resizebox{\textwidth}{!}{%
    \begin{tabular}{c|r|r|r|r|r}
      \multicolumn{1}{l|}{} &
        \multicolumn{1}{c|}{\textbf{ResNet-101}} &
        \multicolumn{1}{c|}{\textbf{ResNet-50}} &
        \multicolumn{1}{c|}{\textbf{Xception}} &
        \multicolumn{1}{c|}{\textbf{VGG-16}} &
        \multicolumn{1}{c}{\textbf{LCNN}} \\ \hline
      \textbf{Error}          & 31,9 & 28,9  & 2,1  & 0,6  & 0,5  \\ \hline
      \textbf{mCE}            & 100  & 105,3 & 94,5 & 63,4 & 48,9 \\ \hline
      Gaussian Noise & 100  & 102   & 140  & 74   & 69   \\
      Shot Noise     & 100  & 98    & 133  & 80   & 63   \\
      Impulse Noise  & 100  & 102   & 137  & 82   & 71   \\
      Speckle Noise  & 100  & 97    & 141  & 78   & 45   \\ \hline
      Defocus Blur   & 100  & 127   & 76   & 80   & 34   \\
      Glass Blur     & 100  & 100   & 40   & 63   & 12   \\
      Motion Blur    & 100  & 115   & 54   & 63   & 19   \\
      Zoom Blur      & 100  & 100   & 40   & 58   & 15   \\
      Gaussian Blur  & 100  & 124   & 75   & 78   & 31   \\ \hline
      Snow           & 100  & 110   & 145  & 57   & 87   \\
      Frost          & 100  & 111   & 142  & 122  & 66   \\
      Fog            & 100  & 108   & 147  & 64   & 114  \\
      Brightness     & 100  & 101   & 124  & 34   & 48   \\
      Spatter        & 100  & 98    & 150  & 63   & 15   \\ \hline
      Contrast       & 100  & 103   & 130  & 66   & 116  \\
      Elastic        & 100  & 94    & 15   & 45   & 5    \\
      JPEG           & 100  & 106   & 15   & 44   & 6    \\
      Pixelate       & 100  & 102   & 19   & 33   & 2    \\
      Saturate       & 100  & 102   & 72   & 21   & 112  \\ \hline
    \end{tabular}%
  }
  \caption{Clean Error, mean Corruption Error (mCE) e per-corruption CE standardized.}
  \label{tab:table_mce}
\end{minipage}

\hfill

\begin{minipage}{0.65\textwidth}
  \centering
  \resizebox{\textwidth}{!}{%
    \begin{tabular}{c|r|r|r|r|r}
      \multicolumn{1}{l|}{} &
        \multicolumn{1}{c|}{\textbf{ResNet-101}} &
        \multicolumn{1}{c|}{\textbf{ResNet-50}} &
        \multicolumn{1}{c|}{\textbf{Xception}} &
        \multicolumn{1}{c|}{\textbf{VGG-16}} &
        \multicolumn{1}{c}{\textbf{LCNN}} \\ \hline
      \textbf{Error}    & 31,9 & 28,9  & 2,1   & 0,6   & 0,5   \\ \hline
      \textbf{Rel. mCE} & 100  & 135,9 & 260,7 & 175,2 & 134,7 \\ \hline
      Gaussian Noise    & 100  & 115   & 287   & 154   & 143   \\
      Shot Noise        & 100  & 106   & 259   & 157   & 124   \\
      Impulse Noise     & 100  & 114   & 274   & 166   & 145   \\
      Speckle Noise     & 100  & 103   & 292   & 162   & 92    \\ \hline
      Defocus Blur      & 100  & 209   & 235   & 259   & 107   \\
      Glass Blur        & 100  & 110   & 77    & 130   & 24    \\
      Motion Blur       & 100  & 173   & 165   & 207   & 59    \\
      Zoom Blur         & 100  & 120   & 114   & 183   & 45    \\
      Gaussian Blur     & 100  & 218   & 273   & 296   & 115   \\ \hline
      Snow              & 100  & 141   & 365   & 144   & 221   \\
      Frost             & 100  & 140   & 332   & 291   & 157   \\
      Fog               & 100  & 133   & 346   & 152   & 275   \\
      Brightness        & 100  & 167   & 870   & 237   & 344   \\
      Spatter           & 100  & 114   & 433   & 184   & 40    \\ \hline
      Contrast          & 100  & 116   & 266   & 137   & 243   \\
      Elastic           & 100  & 103   & 36    & 148   & 12    \\
      JPEG              & 100  & 139   & 34    & 135   & 15    \\
      Pixelate          & 100  & 131   & 53    & 115   & 2     \\
      Saturate          & 100  & 130   & 242   & 72    & 396   \\ \hline
    \end{tabular}%
  }
  \caption{Relative mean Corruption Error (Rel. mCE) e per-corruption Rel. CE standardized.}
  \label{tab:relative_mce}
\end{minipage}
\end{table}

\subsection{Clean Accuracy vs.\ Corruption Robustness}
\begin{figure}[tbp]
  \centering
  \includegraphics[width=0.8\textwidth]{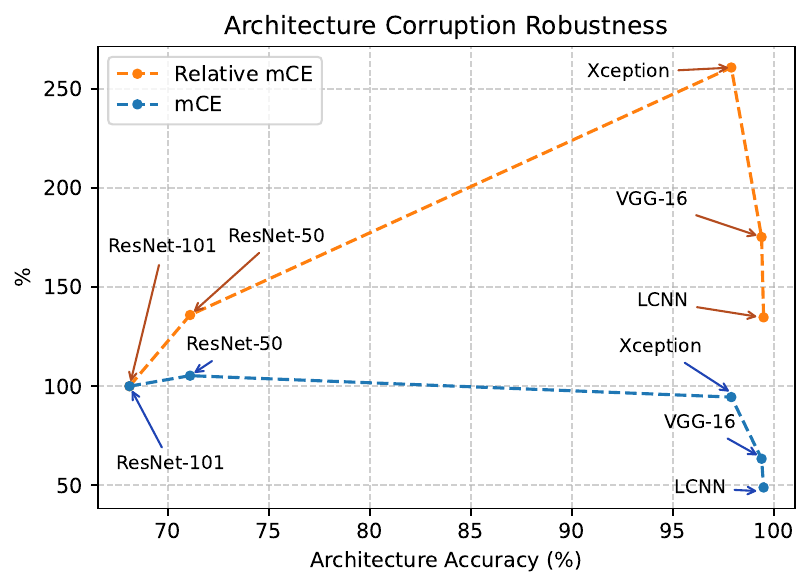}
  \caption{Architecture Accuracy in clean data (MangoLeafDB) vs.\ mCE (blue) and relative mCE (orange) for each architecture.}
  \label{fig:robustness_accuracy}
\end{figure}

To quantify the trade-off between baseline performance and robustness, we plot the clean test precision of each model against its mean corruption error (mCE) and relative mCE (rCE) in Figure~\ref{fig:robustness_accuracy}. 
Several insights emerge from this analysis.
LCNN exhibits distinct Pareto optimality, with a maximum clean accuracy of 99.5\% and an mCE of 48.9, placing it at the forefront of the Pareto front. 
This shows that problem-specific architectures can achieve superior accuracy and robustness simultaneously.

Comparison of ResNet-101 with ResNet-50 shows that while the deeper model slightly improves clean accuracy (68.1\% compared to 71.1\%), it does not significantly enhance robustness. 
Both models have a similar mCE (approximately 100), but ResNet-50 shows a superior relative mCE, indicating that increased depth may offer limited benefits here.

Xception and VGG-16 highlight the equilibrium between model complexity and robustness. 
Xception, characterized by complex layers, achieves a cleaning accuracy of 97.9\% yet presents a higher mCE of 94.5. 
In contrast, VGG-16, with simpler layers, reaches a higher cleaning accuracy of 99.4\% and improved robustness with an mCE of 63.4. 
This emphasizes the intricate balance between performance and resilience to corruption in the model architecture.


In summary, the findings reveal that, for the problem considered in this work, although deeper architectures demonstrate certain robustness, tailored lightweight networks like LCNN and straightforward layer models such as VGG-16 excel in handling corruption without compromising on top-tier cleaning accuracy.

Contextualizing the nature of our comparisons is crucial. This evaluation seeks not to identify a single architecture as the superior choice through statistical means, but to deliver a thorough analysis of how various models handle diverse corruptions. Metrics like F1-score degradation and mCE offer a concise overview of robustness profiles, emphasizing performance disparities vital for practical implementation.

\section{Conclusion}
\label{sec:conclusion}

This research introduced a comprehensive framework to assess the robustness of convolutional neural networks (CNNs) specifically designed to diagnose diseases in mango leaves. 
To facilitate this evaluation, we developed the MangoLeafDB-C dataset, which includes a comprehensive array of 19 different types of corruption manifested at five distinct levels of severity. 
This enables us to emulate diverse conditions that are likely to be encountered in practical real-world applications.
Through these simulations, we conducted a thorough examination of the stability and resilience of five distinct CNN architectures.

The experimental results revealed that the lightweight and specialized LCNN architecture outperformed deeper and more complex models like ResNet-101 and Xception in both clean and corrupted scenarios. 
Notably, LCNN achieved the lowest mean Corruption Error (mCE), maintaining robust performance under distortions such as Defocus Blur, Motion Blur, and various noise-based corruptions. 
In contrast, modern high-capacity models, although accurate under ideal conditions, suffered performance degradation when exposed to corrupted input.

These findings highlight the need to incorporate robustness assessments in the development and deployment of intelligent systems for agriculture. Such robustness evaluation is particularly crucial for regions characterized by limited technological infrastructure and computational resources, where system reliability and efficiency directly impact real-world applicability. 
Lightweight and specialized models, such as LCNN, offer promising solutions in these scenarios, providing not only computational efficiency but also reliability under adverse conditions.

This study has limitations that warrant exploration in future research. 
Firstly, it did not include a comparison between field images and algorithmically generated ones. 
Examining model robustness across domains could be improved, especially regarding performance shifts from controlled datasets to varied real-world environments. 
Secondly, there is substantial scope to investigate sophisticated methods for boosting robustness, such as adversarial training and the creation of noise-resistant loss functions. Lastly, the lack of formal statistical tests to rigorously validate differences between models and corruption suggests that introducing such assessments could enhance future analyses. Nevertheless, we believe our comparative study sufficiently aids researchers in selecting architectures for specific challenging conditions and establishes a foundation for more statistically nuanced future investigations.

%
%
%
\bibliographystyle{splncs04}
\section*{Acknowledgements}
This study was financially supported by Alagoas State Research Support Foundation (FAPEAL) and National Council for Scientific and Technological Development (CNPq) project grant 404825/2023-0.

\bibliography{bib}

@dataset{ali2022mangoleafbd,
  author       = {Ali, Sawkat and Ibrahim, Muhammad and Ahmed, Sarder Iftekhar and Nadim, Md. and Mizanur, Mizanur Rahman and Shejunti, Maria Mehjabin and Jabid, Taskeed},
  title        = {MangoLeafBD Dataset},
  year         = {2022},
  publisher    = {Mendeley Data},
  version      = {V1},
  doi          = {10.17632/hxsnvwty3r.1},
  url          = {https://data.mendeley.com/datasets/hxsnvwty3r/1}
}

@INPROCEEDINGS{detect2023mangodisease,
  author={Mahbub, Nosin Ibna and Naznin, Feroza and Hasan, Md Imran and Shifat, Syed Mahfuzur Rahman and Hossain, Md. Alamgir and Islam, Md Zahidul},
  booktitle={2023 International Conference on Electrical, Computer and Communication Engineering (ECCE)}, 
  title={Detect Bangladeshi Mango Leaf Diseases Using Lightweight Convolutional Neural Network}, 
  year={2023},
  volume={},
  number={},
  pages={1-6},
  organization={IEEE},
  doi={10.1109/ECCE57851.2023.10101648}}

@article{modern2016agriculture,
author = {Rehman, Abdul and Jingdong, Luan and Khatoon, Rafia and Hussain, Imran},
year = {2016},
month = {02},
pages = {284-288},
title = {Modern Agricultural Technology Adoption its Importance, Role and Usage for the Improvement of Agriculture},
volume = {16},
journal = {American-Eurasian Journal of Agricultural \& Environmental Sciences},
doi = {10.5829/idosi.aejaes.2016.16.2.12840}
}

@article{ml2021agriculturereview,
author = {Mirani, Azeem and Memon, Engr Dr Muhammad Suleman and Chohan, Rozina and Wagan, Asif and Qabulio, Mumtaz},
year = {2021},
month = {05},
pages = {229-234},
journal={LUME},
title = {Machine Learning In Agriculture: A Review},
volume = {10}
}

@article{mangotree2021,  title={Flowering induction in mango tree: updates, perspectives and options for organic agriculture},  volume={51},  ISSN={1983-4063},  url={https://doi.org/10.1590/1983-40632021v5168175},  DOI={10.1590/1983-40632021v5168175},  journal={Pesquisa Agropecuária Tropical},  publisher={Escola de Agronomia/UFG},  author={Prates, Adrielle Rodrigues and Züge, Patrícia Graosque Ulguim and Leonel, Sarita and Souza, Jackson Mirellys Azevêdo and Ávila, Jorgiani de},  year={2021},  pages={e68175}}

@Misc{fao2020,
  author = {FOOD AND AGRICULTURE ORGANIZATION OF THE UNITED NATIONS (FAO)},
  note   = {publisher: FAO},
  title  = {Faostat},
  year   = {2020},
  url    = {http://www.fao.org/faostat/en/\#data/QC},
}

@incollection{Citaristi2022fao,
  title={Specialized agencies and related organizations within the un system: Food and agriculture organization of the united nations—fao},
  author={Citaristi, Ileana},
  booktitle={The Europa Directory of International Organizations 2022},
  pages={307--315},
  year={2022},
  publisher={Routledge}
}

@INPROCEEDINGS{mango-2024-tflite,
  author={Patel, Aastha and N, Ravikumar R and Betgeri, Santushti and Singhal, Shilpa and Singh, Sushil Kumar and Takodara, Mitul},
  booktitle={2024 5th International Conference for Emerging Technology (INCET)}, 
  title={Utilizing TFLite and Machine Learning for the Early Detection of Mango Leaf Disease: An Automated Flutter Application}, 
  year={2024},
  volume={},
  number={},
  pages={1-6},
  doi={10.1109/INCET61516.2024.10593509}}

@inproceedings{
hendrycks2018benchmarking,
title={Benchmarking Neural Network Robustness to Common Corruptions and Perturbations},
author={Dan Hendrycks and Thomas Dietterich},
booktitle={International Conference on Learning Representations},
year={2019},
url={https://openreview.net/forum?id=HJz6tiCqYm},
}

@inproceedings{
croce2021robustbench,
title={RobustBench: a standardized adversarial robustness benchmark},
author={Francesco Croce and Maksym Andriushchenko and Vikash Sehwag and Edoardo Debenedetti and Nicolas Flammarion and Mung Chiang and Prateek Mittal and Matthias Hein},
booktitle={Thirty-fifth Conference on Neural Information Processing Systems Datasets and Benchmarks Track (Round 2)},
year={2021},
url={https://openreview.net/forum?id=SSKZPJCt7B}
}

@InProceedings{pmlr-v119-croce20b,
  title = 	 {Reliable evaluation of adversarial robustness with an ensemble of diverse parameter-free attacks},
  author =       {Croce, Francesco and Hein, Matthias},
  booktitle = 	 {Proceedings of the 37th International Conference on Machine Learning},
  pages = 	 {2206--2216},
  year = 	 {2020},
  editor = 	 {III, Hal Daumé and Singh, Aarti},
  volume = 	 {119},
  series = 	 {Proceedings of Machine Learning Research},
  month = 	 {13--18 Jul},
  publisher =    {PMLR},
  url = 	 {https://proceedings.mlr.press/v119/croce20b.html}
}

@InProceedings{Hendrycks_2021_ICCV,
    author    = {Hendrycks, Dan and Basart, Steven and Mu, Norman and Kadavath, Saurav and Wang, Frank and Dorundo, Evan and Desai, Rahul and Zhu, Tyler and Parajuli, Samyak and Guo, Mike and Song, Dawn and Steinhardt, Jacob and Gilmer, Justin},
    title     = {The Many Faces of Robustness: A Critical Analysis of Out-of-Distribution Generalization},
    booktitle = {Proceedings of the IEEE/CVF International Conference on Computer Vision (ICCV)},
    month     = {October},
    year      = {2021},
    pages     = {8340-8349}
}

@article{trinh2024improvingrobustnesscorruptionsmultiplicative,
  title={Improving robustness to corruptions with multiplicative weight perturbations},
  author={Trinh, Quoc Trung and Heinonen, Markus and Acerbi, Luigi and Kaski, Samuel},
  journal={Advances in Neural Information Processing Systems},
  volume={37},
  pages={35492--35516},
  year={2024}
}

@misc{
	michaelis2019whenwinteriscoming,
	author = {Claudio Michaelis and Benjamin Mitzkus and Robert Geirhos and Evgenia Rusak and Oliver Bringmann and Alexander S. Ecker and Matthias Bethge and Wieland Brendel},
	title = {Benchmarking Robustness in Object Detection: Autonomous Driving when Winter is Coming},
	year = {2019},
}
\end{document}